\documentclass[10pt,twocolumn,letterpaper]{article}

\usepackage[T1]{fontenc}
\usepackage[utf8]{inputenc}
\usepackage{amsmath}
\usepackage{newtxtext,newtxmath}
\usepackage[letterpaper,text={6.875in,8.875in},top=1in,columnsep=0.3125in,hcentering]{geometry}
\usepackage{graphicx}
\usepackage{booktabs}
\usepackage{tabularx}
\usepackage{multirow}
\usepackage{siunitx}
\usepackage{xcolor}
\usepackage{algorithm}
\usepackage{algpseudocode}
\usepackage{titlesec}
\usepackage[font=small,labelsep=period]{caption}
\usepackage[numbers,square,sort&compress]{natbib}
\definecolor{linkblue}{rgb}{0.21,0.49,0.74}
\usepackage[breaklinks,colorlinks,citecolor=linkblue,linkcolor=linkblue,urlcolor=linkblue]{hyperref}

\newcolumntype{L}{>{\raggedright\arraybackslash}X}
\newcommand{\stack}[2]{\begin{tabular}[b]{@{}c@{}}#1\\#2\end{tabular}}

\titleformat{\section}{\large\bfseries}{\thesection.}{0.5em}{}
\titlespacing*{\section}{0pt}{10pt plus 2pt minus 2pt}{6pt plus 1pt}
\titleformat{\subsection}{\normalsize\bfseries}{\thesubsection.}{0.5em}{}
\titlespacing*{\subsection}{0pt}{8pt plus 2pt minus 2pt}{4pt plus 1pt}
\titleformat{\paragraph}[runin]{\normalsize\bfseries}{}{0pt}{}[.]
\titlespacing*{\paragraph}{0pt}{4pt plus 1pt}{0.5em}

\renewenvironment{abstract}{\begin{center}\large\bfseries Abstract\end{center}\vspace{-0.6ex}\itshape}{\par\vspace{1.2ex}}

\begin{document}

\twocolumn[{%
\begin{center}
{\Large\bfseries PruneCheck: Closed-Loop Selection for Safe Pruning of\\[2pt] Vision--Language--Action Policies\par}
\vspace{1.4em}
{\large Duy Tran Thanh\qquad Thien-Phuc Doan\qquad Long Nguyen-Vu\qquad Ngo Tan Vu Khanh\par}
\end{center}
\vspace{1.2em}
}]

\begin{abstract}
Vision--language--action (VLA) policies achieve strong results in robot manipulation. Their size makes deployment on robot hardware costly, so they are pruned. Current methods select the components to remove with importance scores borrowed from language-model pruning. These scores are computed without running the policy. They can select components that the closed-loop policy cannot run without. One aggregate success rate does not reveal this error. We propose PruneCheck, a closed-loop pruning framework that selects the components to keep by paired ablation instead of by score. Cheap scores only rank the candidates. Exact paired tests on identical initial states, registered in advance, decide which candidate is kept. On eight public VLA-Adapter checkpoints, PruneCheck reduces 24 gated fusion connections to one while preserving task success on all eight. The learned gate selects a sufficient connection on five checkpoints. Block Influence, the score of ShortGPT and EfficientVLA, does so on two. Its choices reduce success from 100\% to 0\% on one task suite. On the shipped checkpoint, the pruned policy reaches 91.8\% success against 93.0\% over 500 episodes. PruneCheck costs 6 to 9 GPU-hours per checkpoint on one GPU.
\end{abstract}

\section{Introduction}\label{sec:intro}
\begin{figure*}[!tp]
\centering
\includegraphics[width=\textwidth]{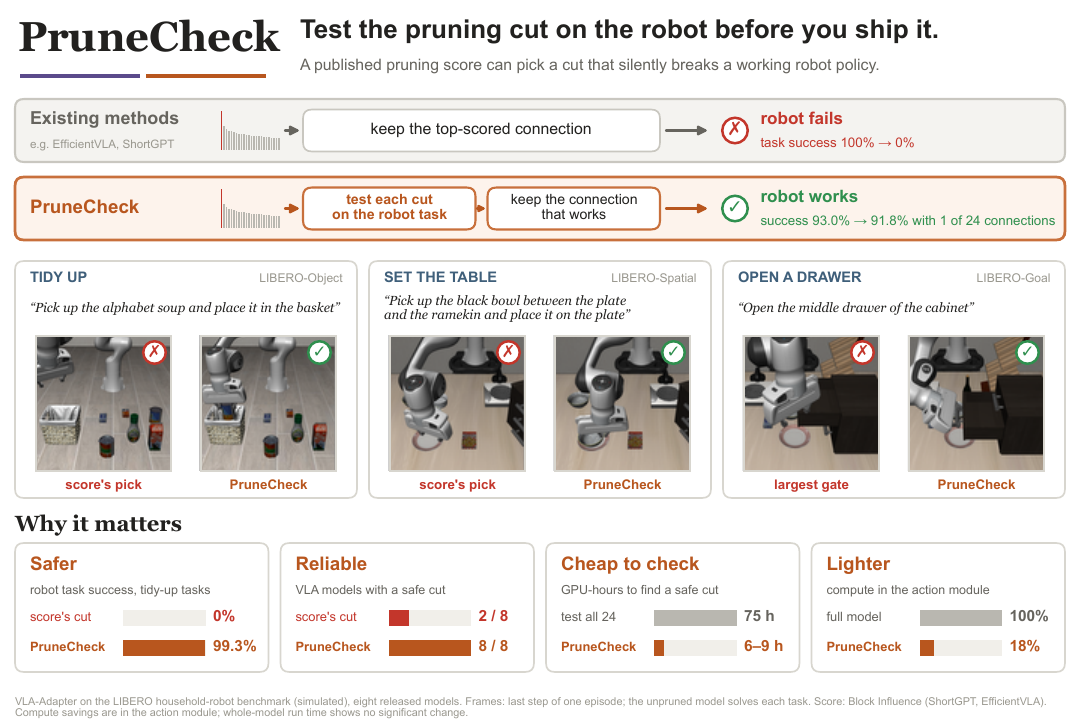}
\caption{\textbf{PruneCheck at a glance.} Top: existing score-based pruning methods keep the connection a score ranks highest. PruneCheck tests each cut on the robot task and keeps the connection that works. Middle: the final frame of one episode under each pick, on three household tasks. Bottom: PruneCheck against the existing choice on task success, reliability across eight released VLA models, the cost of the check and the compute of the action module. Whole-model run time shows no significant change.}
\label{fig:interface}
\end{figure*}

Vision--language--action (VLA) policies have become a leading approach to robot control \citep{kim2025openvlaoft,intelligence2025pi05,bjorck2025gr00t}. A VLA policy turns camera images, a language instruction and the robot's state into motor commands. A large pretrained vision--language model (VLM), the backbone, processes the images and the instruction. An action module turns the backbone's features into motor commands. A VLA policy is evaluated in closed loop. It controls a simulated robot through each whole task, and every action changes the next image it sees. The score is the success rate on a benchmark such as LIBERO \citep{liu2023libero}.

Designs differ in how the action module reads the backbone. OpenVLA \citep{kim2024openvla} predicts actions as tokens of the backbone itself. OpenVLA-OFT \citep{kim2025openvlaoft} attaches an action head to the last backbone layer. GR00T N1 \citep{bjorck2025gr00t} reads one intermediate layer. $\pi_{0.5}$ \citep{intelligence2025pi05}, SmolVLA \citep{shukor2025smolvla} and VLA-Adapter \citep{wang2026vlaadapter} use a separate action module that attends to backbone features at several layers. VLA-Adapter adds a learned gate to each such connection. Such gates come from earlier work on pretrained networks: ReZero \citep{bachlechner2021rezero}, LayerScale \citep{touvron2021cait} and Flamingo \citep{alayrac2022flamingo}. A gate initialized at or near zero leaves the pretrained network almost unchanged at the start of training. A trained gate is therefore easy to read as a measure of use. Whether each modality is used is a long-standing question in multimodal fusion \citep{baltrusaitis2019multimodal,wang2020multimodal}. It is rarely asked of individual gated connections.

These policies must run on the limited compute of a robot, so a growing line of work prunes them. EfficientVLA \citep{yang2025efficientvla} removes backbone layers by Block Influence. This is the layer score of ShortGPT \citep{men2025shortgpt}. ShortGPT removes the layers whose output is most similar to their input \citep{gromov2024unreasonable,men2025shortgpt}. MoLe-VLA \citep{zhang2026molevla} trains a router that skips layers. DeeR-VLA \citep{yue2024deervla} stops the backbone early. AutoPruner \citep{lin2026autopruner} selects backbone layers with an open-loop probe and verifies the pruned policy in closed loop. Many of these scores come from language-model pruning, where most attention heads of a trained transformer were shown to be removable \citep{michel2019sixteen,voita2019analyzing}. These methods report one success rate for the pruned policy. None measures, for each fusion connection, whether the policy needs it. None tests a published score against closed-loop ablation.

A score is computed from forward passes, without rollouts. It is a prediction of what the policy can lose. The property at stake is the faithfulness of the score: whether it predicts the effect of removing the component. Attention weights often disagree with other measures of importance \citep{jain2019attention}. For language models, cosine similarity is a poor proxy for the loss caused by removing a layer \citep{hinostroza2026rethinking}. Pruning results vary across models and datasets \citep{blalock2020state}. The best backbone layer to read in a VLA does not transfer across backbones \citep{liu2026layermystery}. One success rate shows that one pruned policy works. It does not show that the score selected the right components. A score that is right on one checkpoint can then fail on the next release of the same model.

Evaluation harnesses such as vla-eval \citep{choi2026vlaeval} measure how well a policy performs. The robustness benchmarks LIBERO-Plus \citep{fei2025liberoplus} and LIBERO-PRO \citep{zhou2025liberopro} find policies brittle to camera, initial state and object changes. Mechanistic studies inject activations \citep{grant2026features} or mask visual tokens layer by layer \citep{luo2026lookbefore}. Injection tests what suffices to drive behaviour. A pruning decision needs the other kind of evidence: what the policy cannot do without. Sanity checks for saliency maps \citep{adebayo2018sanity} set the pattern for such evidence: a test that any score must pass.

We argue that current pruning practice misses a crucial point: a pruning decision is a claim about the closed-loop policy, so it must be tested in closed loop. Building on this principle, we propose PruneCheck, a closed-loop pruning framework that selects the components to keep by paired ablation (Fig.~\ref{fig:interface}). Our design is guided by four principles.

\paragraph{Scores rank, ablation decides} Cheap scores order the candidates, so that the most likely ones are rolled out first. A candidate is kept only when its closed-loop ablation supports it.

\paragraph{Keep-one and drop-all arms} The keep-one arm keeps a single candidate and removes the rest. It shows whether that candidate suffices. The drop-all arm removes every candidate. It shows whether the components matter at all.

\paragraph{Paired exact tests} All arms run on the same initial states. An exact McNemar test, a test for paired success and failure outcomes, compares each arm with the unmodified policy. Pairing removes the variation between initial states from the comparison.

\paragraph{Registration before every run} Each comparison fixes its prediction, its margin and its refutation before it runs. A selection is therefore a tested decision, not a reading of the results after the fact.

We apply PruneCheck to the fusion interface of VLA-Adapter. Its action decoder reads the backbone through 24 connections. Each connection links one decoder block to one backbone layer. A learned scalar, the gate, scales what the block receives. VLA-Adapter releases eight checkpoints, which we call heads. They cover two releases and the four LIBERO suites. PruneCheck reduces the interface from 24 connections to one on all eight heads. Keeping the selected connection alone changes success by $-2.0$ to $+3.3$ percentage points (Fig.~\ref{fig:pruning}). We compare PruneCheck with the selection criteria a practitioner would use instead (Table~\ref{tab:published}). The largest gate selects a sufficient connection on five heads. Block Influence does so on two. Its choices reduce success from 96.0\% to 14.7\% on LIBERO-Spatial. They reduce it from 100\% to 0\% on LIBERO-Object. On the shipped LIBERO-Long head, the pruned policy reaches 91.8\% success against 93.0\% over 500 episodes. Its action decoder performs 82.2\% fewer multiply-accumulates. PruneCheck costs 6 to 9 GPU-hours per head on one Tesla T4. On the image captioner OpenFlamingo-3B \citep{awadalla2023openflamingo}, PruneCheck shows the opposite structure. The model's function is spread over many blocks. Its four largest gates recover only 16\% of the image's effect.

Our key contributions are as follows:
\begin{itemize}
\item \textbf{Closed-loop pruning framework.} PruneCheck formulates a pruning decision as a testable claim about task success under a component mask. It selects what to keep by paired ablation and runs on one commodity GPU.
\item \textbf{Comparison with published selection criteria.} In a registered comparison, PruneCheck selects a sufficient connection on eight of eight heads. The learned gate and three forms of Block Influence succeed on two to six.
\item \textbf{A 24-fold smaller fusion interface.} One connection suffices on every head. On the first-release heads, a rank collapse in the backbone explains this.
\item \textbf{A reporting standard for pruning studies.} We recommend that every pruning study report keep-one and drop-all arms next to its pruned success rate.
\end{itemize}
Through these contributions, PruneCheck turns the selection step of VLA pruning from an untested prediction into a measured decision.

\section{PruneCheck Framework}\label{sec:method}

\subsection{Target policy}
\label{sec:system}

VLA-Adapter \citep{wang2026vlaadapter} pairs a Qwen2.5-0.5B language-model backbone \citep{qwen2024qwen25} with a 24-block action decoder (Fig.~\ref{fig:interface}). Two image encoders, DINOv2 \citep{oquab2024dinov2} and SigLIP \citep{zhai2023siglip}, turn the camera images into patch tokens for the backbone. In one softmax, decoder block $l$ attends to its own action tokens, to proprioceptive tokens that encode the robot's state, and to the 512 patch tokens that leave backbone layer $l+1$. Blocks are numbered from 0 and layers from 1, so block 23 reads the last layer. A learned gate, $\tanh(g_l)$, multiplies the attention scores of the patch tokens. We call each such read a \emph{connection}. The authors release eight checkpoints, which we call heads. Four are first-release heads, one per LIBERO suite: Spatial, Object, Goal and Long. Four are heads of a later revision, called Pro, on the same suites. Each head has its own adapted backbone and decoder.

We chose VLA-Adapter for three reasons. It is a strong policy: its authors report average LIBERO success of 97.3\% for the first release and 98.5\% for Pro. It places a learned gate on every layer-wise connection, the design in which reading the gate as importance is most tempting. It releases checkpoints for all four suites in two releases. To our knowledge, no other VLA with learned per-layer gates releases such a set.

Removing a connection means deleting block $l$'s patch tokens from the keys of its softmax. Zeroing the gate would not remove the connection. With a zero gate, every patch token still receives the same non-zero attention weight, so the block receives the mean of the patch tokens. With every connection kept, the removal code reproduces the original forward pass bit for bit.

\subsection{Problem formulation}
\label{sec:needed}

Let a policy have $L$ connections, indexed $l = 0, \dots, L-1$. Here $L = 24$. A mask $m \in \{0,1\}^{L}$ keeps connection $l$ when $m_l = 1$ and removes it when $m_l = 0$. Let $y_i(m) \in \{0,1\}$ be the outcome of the episode that starts from initial state $i$ under mask $m$. One run of the policy from an initial state to the end of the task is an episode, also called a rollout. An episode succeeds, $y_i = 1$, when LIBERO's own success predicate fires within the suite's standard episode limit. The success rate over $n$ initial states is
\begin{equation}
S(m) = \frac{1}{n} \sum_{i=1}^{n} y_i(m).
\label{eq:success}
\end{equation}
The unmodified head is the mask $m = \mathbf{1}$. An importance score $s$ assigns a number $s(l)$ to each connection without any rollout. Pruning to one connection by the score keeps $l^{*} = \arg\max_l s(l)$. We write $e_{l}$ for the mask that keeps only connection $l$. The pruning decision is then the claim $S(e_{l^{*}}) \approx S(\mathbf{1})$. The score cannot verify this claim, because it is not computed from any outcome $y_i$. PruneCheck measures the paired difference
\begin{equation}
\Delta(m) = S(m) - S(\mathbf{1}) = \frac{1}{n} \sum_{i=1}^{n} \bigl( y_i(m) - y_i(\mathbf{1}) \bigr)
\label{eq:delta}
\end{equation}
on the same initial states for both masks.

We measure $\Delta$ under three test conditions, called arms. All arms run on the same $n = 150$ initial states. These are LIBERO's fixed states, 15 for each of the 10 tasks of a suite. The first arm is the unmodified head, $m = \mathbf{1}$. The second, \emph{keep one}, uses $m = e_l$ and removes every connection but one. The third, \emph{drop all}, uses $m = \mathbf{0}$ and removes every connection. Keep one shows whether a single connection suffices. Drop all shows whether the interface matters at all. Neither is informative alone.

Each arm is compared with the unmodified head by an exact McNemar test \citep{mcnemar1947note}. Let $b$ be the number of states solved only by the unmodified head, and $c$ the number solved only by the arm. Under the hypothesis of equal success, $b$ follows a binomial distribution with $b + c$ trials and probability $1/2$. The test gives the two-sided $p$-value of the observed $b$. It uses only the states on which the two arms disagree. A change is significant when this $p$ is below 0.05.

Differences in success rate are given in percentage points (points).

\subsection{Cheap screens}
\label{sec:screens}

A screen is a score that is computed without any rollout. We test five screens that a practitioner would use instead of rollouts. Each takes minutes per checkpoint.

\textbf{Gate census.} The magnitude $|\tanh g_l|$ of each gate. The connection with the largest gate is taken as the one to keep.

\textbf{Attention selectivity.} This screen is $1-\bar{H}$, with $\bar{H}$ the normalized entropy of the attention weights over a connection's 512 tokens. A connection that averages its tokens has selectivity near zero.

\textbf{Input rank.} This screen is the participation ratio of the tokens a connection receives,
\begin{equation}
\mathrm{PR} = \frac{\left(\sum_i \lambda_i\right)^2}{\sum_i \lambda_i^2},
\label{eq:pr}
\end{equation}
with $\lambda_i$ the eigenvalues of their covariance. PR counts the effective number of directions in which the tokens vary.

\textbf{Offline recovery.} This screen uses the action loss on demonstration frames with all connections ($L_{\mathrm{full}}$), with only the largest-gate connection ($L_{\mathrm{keep}}$) and with none ($L_{\mathrm{none}}$):
\begin{equation}
R = \frac{L_{\mathrm{none}} - L_{\mathrm{keep}}}{L_{\mathrm{none}} - L_{\mathrm{full}}}.
\label{eq:recovery}
\end{equation}
A value near one means the kept connection restores the full loss. A value near zero means it restores nothing.

\textbf{Block Influence.} For layer $j$, this published score is
\begin{equation}
\mathrm{BI}_j = 1 - \mathbb{E}\left[\cos\left(x_j^{\mathrm{in}}, x_j^{\mathrm{out}}\right)\right],
\label{eq:bi}
\end{equation}
where the cosine compares the layer's input and output hidden states over calibration samples and token positions. A layer that changes its input little has a low score and is treated as redundant. To keep one connection, we keep the one with the highest score. We compute the score in three forms. \textbf{C1} uses the backbone layer the connection reads, over all token positions, and is the published score. \textbf{C2} uses the same layer over the 512 positions the connection reads. \textbf{C3} uses the decoder block itself, over its action latents. These forms apply the published score to the fusion interface. They are not reproductions of the two pruning methods, which remove backbone layers.

\subsection{PruneCheck protocol}
\label{sec:protocol}

Algorithm~\ref{alg:prunecheck} states PruneCheck. Screens rank what to roll out first and never decide. The paired ablation decides. When keeping the largest gate is no better than dropping everything, the next candidate is ablated, the deepest block first. We call this the fallback rule. The registered comparisons of this paper use a margin $\delta = 8$ points and a level $\alpha = 0.05$. Each arm costs about 3 GPU-hours on a Tesla T4. A verdict on one candidate therefore costs about 6 GPU-hours, and each further candidate adds about 3.

\begin{algorithm}[t]
\caption{PruneCheck for one checkpoint}
\label{alg:prunecheck}
\small
\begin{algorithmic}[1]
\Require checkpoint with $L$ connections, screens $s_1, \dots, s_K$, $n$ paired initial states, margin $\delta$, level $\alpha$
\State Compute every screen on every connection \Comment{forward passes only}
\State $Q \gets$ ordered candidates: the largest-gate connection, the deepest connection, then the other screen picks
\State Run the unmodified arm $m = \mathbf{1}$ and the drop-all arm $m = \mathbf{0}$ on the $n$ states
\If{$\Delta(\mathbf{0})$ is not significantly below zero}
  \State \Return the interface is not needed
\EndIf
\For{each connection $l$ in $Q$}
  \State Run the keep-one arm $m = e_l$ on the same $n$ states
  \State Compute $\Delta(e_l)$ and the exact McNemar $p$ against the unmodified arm
  \If{$p \ge \alpha$ and $\Delta(e_l) \ge -\delta$}
    \State \Return prune to connection $l$
  \EndIf
\EndFor
\State \Return no single candidate suffices
\end{algorithmic}
\end{algorithm}

\subsection{Experimental record}
\label{sec:record}

Every closed-loop experiment was registered before it ran. Each registration fixed a numeric prediction, the result that would refute it, and instrument conditions that check the measurement itself. A failed condition is recorded as ``not tested'', never as a success. Rollouts use LIBERO's fixed initial states and $256\times256$ images. They run in 16-bit floating point (fp16) on Tesla T4 GPUs. The comparison with published criteria uses a second software environment: the VLA-Adapter authors' released code with their package versions and software rendering. PruneCheck is implemented as a Python package with resumable paired runs, so it runs on one commodity GPU. Its VLA-Adapter adapter reproduced all 450 recorded closed-loop outcomes of three arms on 150 states. Supplementary Information lists every registered outcome, including the validation runs of the package.

\section{Experiments}\label{sec:results}

Section~\ref{sec:method} defined the policy, the arms, the screens and the tests. We first report the closed-loop verdicts on the eight heads. We then compare PruneCheck with cheap screens and with Block Influence, test its fallback rule prospectively and prune the shipped checkpoint. The last two subsections explain the single connection and apply PruneCheck to two other models.

\subsection{A single sufficient connection and a misleading gate}
\label{sec:suffices}

\begin{figure*}[!tp]
\centering
\includegraphics[width=0.72\textwidth]{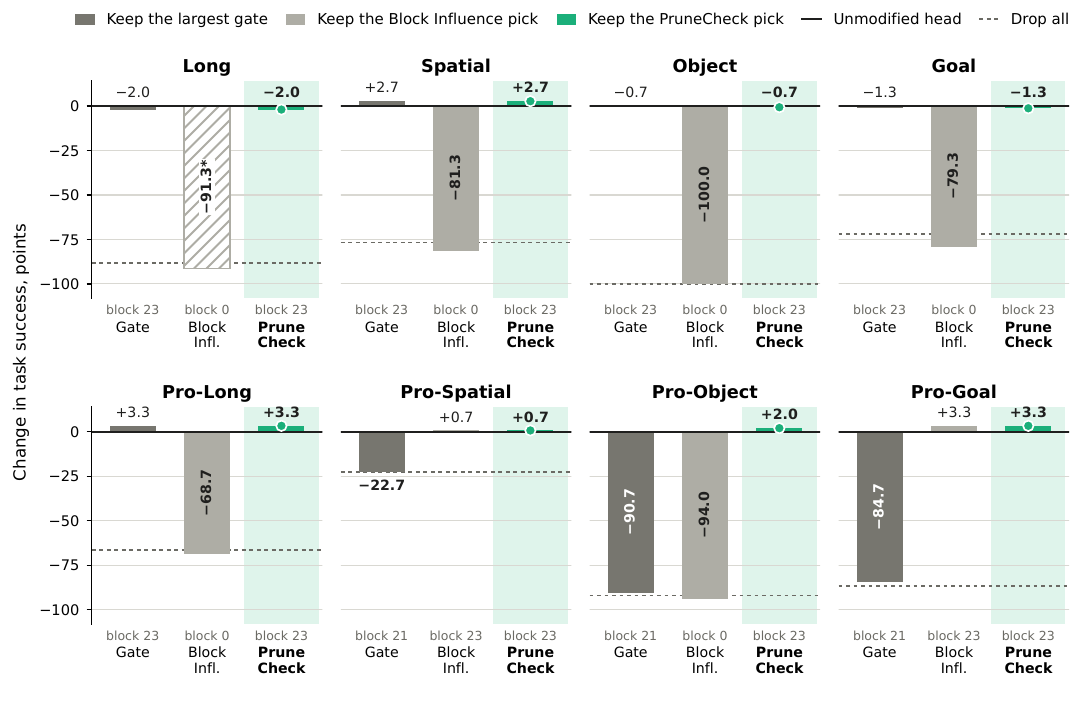}
\caption{\textbf{What each selection rule costs when a head is pruned to one connection.} Each bar gives the paired change in closed-loop success against the unmodified head when only the selected block is kept, over 150 states. The dashed line marks the cost of removing all 24 connections. The PruneCheck pick matches the unmodified head on all eight heads. $^{*}$Exploratory, because the reproduction condition failed on this head. Exact tests are in Tables~\ref{tab:screens} and~\ref{tab:published} and in the text.}
\label{fig:pruning}
\end{figure*}

On the four first-release heads, we kept only block 23, the deepest connection. Success changes by $-2.0$ points on Long, $+2.7$ on Spatial, $-0.7$ on Object and $-1.3$ on Goal. No change is significant: every exact McNemar $p$ is at least 0.42. Removing all 24 connections costs 88.0, 76.7, 100.0 and 72.0 points on the same heads (Fig.~\ref{fig:pruning}, Table~\ref{tab:screens}). The interface is therefore needed, and one connection carries it. On these heads, block 23 also carries the largest gate, so the gate is right.

The gate is not right on the next release. Pro-Long behaves like the first-release heads: keeping block 23 changes success by $+3.3$ points. On Pro-Spatial, Pro-Goal and Pro-Object, the largest gate sits at block 21. Keeping that connection alone is no better than keeping none. On Pro-Object, the two arms lose 90.7 and 92.0 points. The interface can thus shrink 24-fold, but the gate would keep the wrong connection on three of eight heads.

\subsection{Failures of the cheap screens}
\label{sec:screensfail}

Table~\ref{tab:screens} compares the screens with the closed-loop verdict. Beyond its ranking, the gate also fails as an absolute reading. On first-release Spatial and Goal, the connection that suffices has a gate of 0.0045 and 0.0016. Both are below 0.01, the threshold we registered for calling a gate open. Attention selectivity is near zero on the same two heads, at $1.0\times10^{-6}$ and $1.0\times10^{-7}$. Their needed connection delivers only the mean of its 512 tokens. Yet removing it together with the other 23 costs 76.7 and 72.0 points. Uniform attention is therefore not evidence of disuse.

Input rank was registered with the prediction $\mathrm{PR} > 2$ at the largest-gate block exactly where that block suffices. Two of eight heads fell on the wrong side. Pro-Spatial and Pro-Goal have rank 8.45 and 5.11 at a block that is worth nothing alone. Offline recovery separated the eight heads correctly. $R$ ranges from 0.974 to 0.997 where the largest gate suffices and from 0.005 to 0.048 where it does not. It is itself an ablation, scored on loss, and it was registered with the verdicts already known. We use it to order rollouts. None of the screens can replace the closed-loop test.

\begin{table*}[!tp]
\caption{\textbf{Cheap screens and closed-loop results on the eight released heads.} The screens are computed without running the policy. The closed-loop columns give the change in task success, in percentage points, against the unmodified head over 150 paired episodes. A negative value is a loss. Input rank is the median participation ratio of the tokens that reach the largest-gate block. Offline recovery is Eq.~\ref{eq:recovery} for that block.}
\label{tab:screens}
\small
\centering
\begin{tabular}{@{}lc cc ccc c@{}}
\toprule
 & & \multicolumn{2}{c}{Screens, no rollouts} & \multicolumn{3}{c}{Change in success when keeping} & \\
\cmidrule(lr){3-4}\cmidrule(lr){5-7}
Head & \stack{Largest-gate}{block} & \stack{Input}{rank} & \stack{Offline}{recovery} & \stack{only the}{largest-gate block} & \stack{only}{block 23} & \stack{no}{connection} & \stack{Largest gate}{suffices?} \\
\midrule
Long        & 23 & 17.63 & 0.993 & $-2.0$  & $-2.0$  & $-88.0$  & yes \\
Spatial     & 23 & 19.35 & 0.987 & $+2.7$  & $+2.7$  & $-76.7$  & yes \\
Object      & 23 & 10.93 & 0.997 & $-0.7$  & $-0.7$  & $-100.0$ & yes \\
Goal        & 23 & 17.94 & 0.981 & $-1.3$  & $-1.3$  & $-72.0$  & yes \\
Pro-Long    & 23 & 26.95 & 0.974 & $+3.3$  & $+3.3$  & $-66.7$  & yes \\
Pro-Object  & 21 &  1.19 & 0.048 & $-90.7$ & $+2.0$  & $-92.0$  & no  \\
Pro-Spatial & 21 &  8.45 & 0.005 & $-22.7$ & $+0.7$  & $-22.7$  & no  \\
Pro-Goal    & 21 &  5.11 & 0.033 & $-84.7$ & $+3.3$  & $-86.7$  & no  \\
\bottomrule
\end{tabular}
\end{table*}

\subsection{Selection criteria from published pruning methods}
\label{sec:published}

A pruning study is more likely to select components with a published score than with the gate. We therefore tested Block Influence (Eq.~\ref{eq:bi}) in three forms on the same eight heads. C1 is the published form, computed on the backbone layer a connection reads. C2 and C3 adapt it to the fusion interface (Sec.~\ref{sec:screens}). The comparison was registered before any score was computed. The connection a criterion selects is its pick. A pick \emph{suffices} if keeping it alone gives an exact McNemar $p \ge 0.05$ and a change no worse than $-8$ points. A pick \emph{fails} if the change is $-20$ points or worse with $p < 0.05$. Picks already tested above take those results. Every other pick was rolled out on the 150 states in the second software environment and paired against a new run of the unmodified head. A registered condition required each new unmodified run to land within 5 successes of the count recorded in the first environment. It held on five heads and failed on Long, with 144 against 138. The Long comparison is therefore reported as exploratory.

\begin{table*}[!tp]
\caption{\textbf{The connection each criterion keeps, and what keeping it alone costs.} Each cell gives the kept block, then in parentheses the change in task success, in percentage points, when only that block is kept (150 paired episodes). Bold marks a failure: a loss of 20 points or more with $p<0.05$. Gate: the largest learned gate. C1: Block Influence as published. C2 and C3: Block Influence adapted to the fusion interface (Sec.~\ref{sec:screens}). PruneCheck: the block selected by closed-loop ablation. $^{*}$Exploratory, because the reproduction check failed on this head.}
\label{tab:published}
\small
\centering
\begin{tabular}{@{}lccccc@{}}
\toprule
Head & Gate & C1 & C2 & C3 & PruneCheck \\
\midrule
Long & 23 ($-2.0$) & 0 ($-91.3$)$^{*}$ & 23 ($-2.0$) & 23 ($-2.0$) & 23 ($-2.0$) \\
Spatial & 23 ($+2.7$) & {\bfseries\boldmath 0 ($-81.3$)} & 23 ($+2.7$) & 23 ($+2.7$) & 23 ($+2.7$) \\
Object & 23 ($-0.7$) & {\bfseries\boldmath 0 ($-100.0$)} & {\bfseries\boldmath 16 ($-99.3$)} & 23 ($-0.7$) & 23 ($-0.7$) \\
Goal & 23 ($-1.3$) & {\bfseries\boldmath 0 ($-79.3$)} & {\bfseries\boldmath 1 ($-76.7$)} & 23 ($-1.3$) & 23 ($-1.3$) \\
Pro-Long & 23 ($+3.3$) & {\bfseries\boldmath 0 ($-68.7$)} & 23 ($+3.3$) & 23 ($+3.3$) & 23 ($+3.3$) \\
Pro-Spatial & {\bfseries\boldmath 21 ($-22.7$)} & 23 ($+0.7$) & 23 ($+0.7$) & {\bfseries\boldmath 20 ($-22.7$)} & 23 ($+0.7$) \\
Pro-Object & {\bfseries\boldmath 21 ($-90.7$)} & {\bfseries\boldmath 0 ($-94.0$)} & {\bfseries\boldmath 16 ($-93.3$)} & 23 ($+2.0$) & 23 ($+2.0$) \\
Pro-Goal & {\bfseries\boldmath 21 ($-84.7$)} & 23 ($+3.3$) & 23 ($+3.3$) & {\bfseries\boldmath 22 ($-92.0$)} & 23 ($+3.3$) \\
\midrule
Heads where the kept block suffices & 5 of 8 & 2 of 8 & 5 of 8 & 6 of 8 & 8 of 8 \\
\bottomrule
\end{tabular}
\end{table*}

C1, the published form, selects block 0 on six heads (Table~\ref{tab:published}). Wherever this choice could be evaluated, it fails. Keeping only block 0 loses 68.7 to 100.0 points on five heads, each with an exact McNemar $p < 10^{-30}$. On Long, the exploratory loss is 91.3 points. C2 moves the pick to block 23 on five heads and fails on the other three. C3 selects block 23 on six heads and fails on Pro-Spatial and Pro-Goal. No criterion is right on every head. In every failure, keeping block 23 matches the unmodified head.

\begin{figure*}[!tp]
\centering
\includegraphics[width=0.72\textwidth]{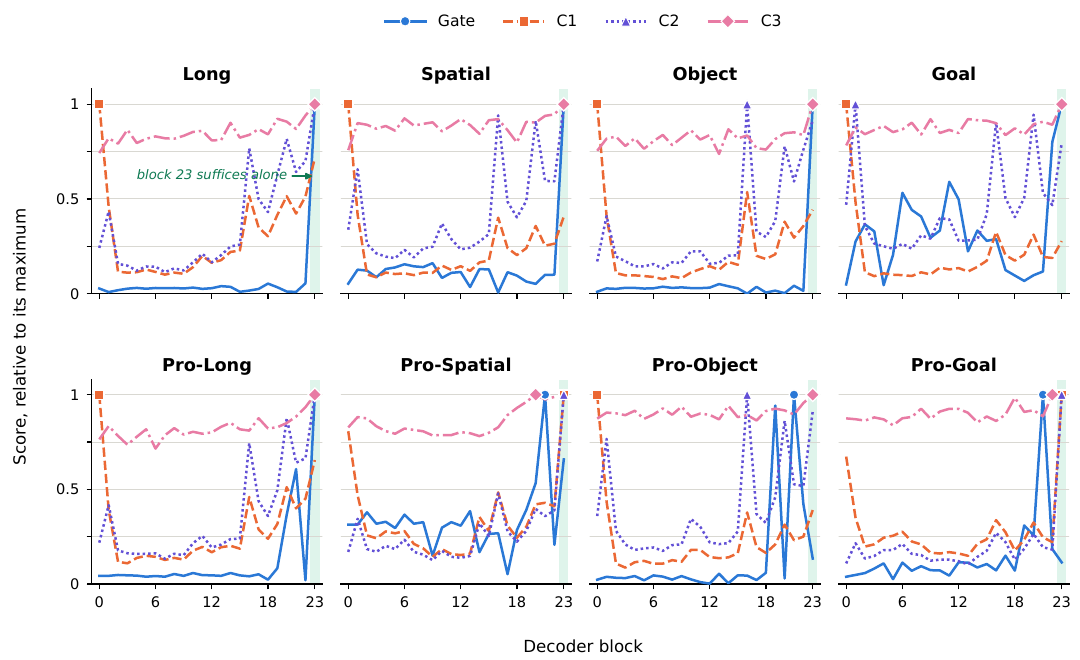}
\caption{\textbf{Score profiles over the 24 connections.} Each curve is one score, relative to its maximum on that head. A marker shows the block the score selects. The shaded band marks block 23, which suffices alone on every head. Gate: $|\tanh g_l|$. C1 to C3: Block Influence in its three forms.}
\label{fig:profiles}
\end{figure*}

Figure~\ref{fig:profiles} shows why the scores disagree with the ablation. C1 peaks at block 0 on six heads and reaches only 0.28 to 0.72 of its maximum at block 23. The gate has one sharp peak, at block 23 on the first release and at block 21 on three Pro heads. C3 is nearly flat, so small differences decide its maximum. Block Influence measures how much a layer changes its input. It does not measure whether the decoder uses what the layer delivers. Each score has a clear maximum on every head. Only the ablation shows whether that maximum is the right connection.

\begin{figure*}[!tp]
\centering
\includegraphics[width=0.62\textwidth]{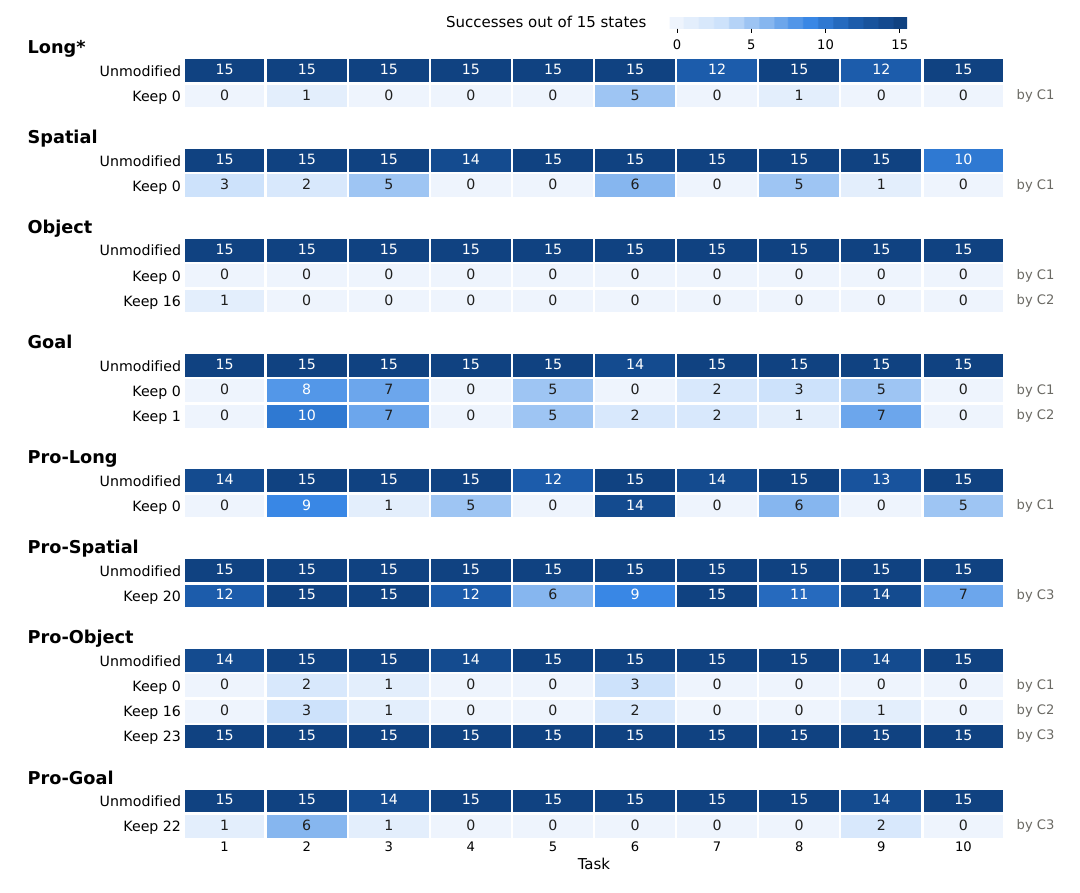}
\caption{\textbf{Success per task in the registered comparison.} Each cell gives the number of solved states out of 15 for one task. Rows are the unmodified head and the keep-one arms run in the second environment. The label at the right names the criterion that selected the block. $^{*}$Exploratory.}
\label{fig:pertask}
\end{figure*}

\begin{figure*}[!tp]
\centering
\includegraphics[width=0.68\textwidth]{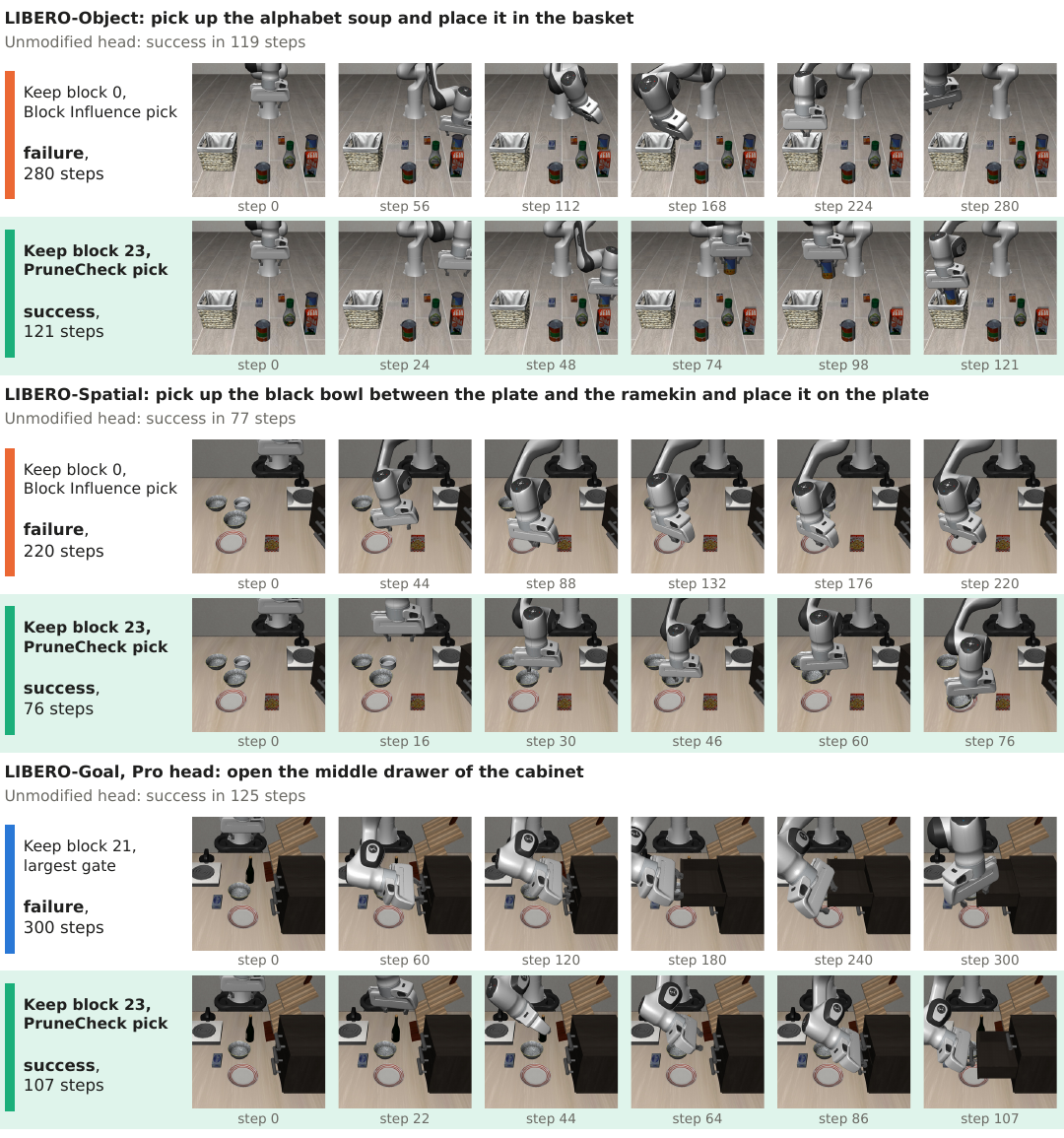}
\caption{\textbf{Three tasks under a failing pick and the PruneCheck pick.} Both rows of a group start from the same initial state. Each header gives the result of the unmodified head on that state. Frames are evenly spaced over each episode and show a high-resolution view of the scene. The policy input is unchanged. The Block Influence pick and the largest gate fail at the step limit. The PruneCheck pick behaves like the unmodified head.}
\label{fig:rollout}
\end{figure*}

The failures are not spread evenly over tasks (Fig.~\ref{fig:pertask}). Keeping block 0 on Pro-Long solves 14 of 15 states of one task and none of four others. Keeping block 20 on Pro-Spatial leaves three tasks at 15 of 15 and drops two to 6 and 7. One aggregate success rate hides this structure. Figure~\ref{fig:rollout} shows single episodes. With only the failing pick kept, the arm moves but the episode ends at the step limit. With only block 23 kept, the policy completes each task in about as many steps as the unmodified head. These episodes are illustrations outside the registered record. On Pro-Goal, all three policies solved the first two states of the task, and the third state is shown.

\subsection{Prospective test of the fallback rule}
\label{sec:prospective}

The fallback rule of Algorithm~\ref{alg:prunecheck} says: when the largest gate fails, ablate the deepest block. We wrote this rule after the Pro failures. We then registered it as a prediction on the two failing heads where block 23 had not been tried. The prediction was a change no worse than $-8$ points, and it held on both heads. On Pro-Spatial, keeping only block 23 solves 150 of 150 states. The unmodified head solves 149 ($+0.7$ points, McNemar $p=1.0$). On Pro-Goal, keeping only block 23 solves 148 of 150. The unmodified head solves 143 ($+3.3$, $p=0.125$). Keeping the largest gate solves only 115 and 16 on these heads. On Pro-Object, block 23 was rolled out as the C3 pick of the comparison with published criteria. It solves 150 of 150 states against 147 ($+2.0$ points, $p=0.25$). One connection therefore suffices on all eight heads, and it is block 23 on each.

\subsection{Pruning the shipped checkpoint}
\label{sec:deploy}

Table~\ref{tab:deploy} collects the registered measurements on the Long head. We pruned the shipped checkpoint at inference to block 23, with weights untouched, and evaluated it on all 500 initial states of the suite. Success moves from 93.0\% to 91.8\%, a change of $-1.2$ points that is within the registered bound of $-4.0$ points. The cut removes 82.2\% of the decoder's multiply-accumulate operations. After brief fine-tuning of the pruned action decoder, 47 GPU-minutes, the difference on 150 paired states is not significant. The saving is in arithmetic, not in time. The backbone dominates the action query, so the median end-to-end saving is 1.8\%, within noise. The case for PruneCheck is therefore that the cut it selects does not harm the policy. The cut the gate suggests can destroy it.

\begin{table*}[!tp]
\caption{\textbf{Pruning the shipped LIBERO-Long head to block 23.} Top: task success. Each pruned policy is compared with the unmodified head on the same initial states, using an exact McNemar test. Bottom: compute on one Tesla T4.}
\label{tab:deploy}
\small
\centering
\begin{tabular}{@{}lcccc@{}}
\toprule
Policy & Episodes & Success & Change (points) & Significant? \\
\midrule
Unmodified head & 500 & 93.0\% & -- & -- \\
Pruned to block 23 & 500 & 91.8\% & $-1.2$ & no ($p=0.41$) \\
\addlinespace
Unmodified head & 150 & 92.0\% & -- & -- \\
Pruned to block 23 & 150 & 90.0\% & $-2.0$ & no ($p=0.58$) \\
Pruned to block 23, then fine-tuned for 4,000 steps & 150 & 92.7\% & $+0.7$ & no \\
All 24 connections removed & 150 & 4.0\% & $-88.0$ & yes ($p<0.001$) \\
\midrule
Compute & & Unmodified & Pruned & Change \\
\midrule
Decoder multiply-accumulates & & -- & -- & 82.2\% fewer \\
Action-decoder call & & 20.5\,ms & 16.5\,ms & 19.6\% faster \\
Whole action query & & 146.9\,ms & 141.3\,ms & 1.8\% faster, within noise \\
\bottomrule
\end{tabular}
\end{table*}

\subsection{Mechanism behind the single connection}
\label{sec:mechanism}

The single connection has a measurable cause. The tokens that the backbone hands to mid-depth connections have a participation ratio near one in all four first-release heads. Over the 20 mid-depth blocks, the per-head median is 1.08 to 1.14. The collapse comes from the backbone. The MLP of its fourth layer writes a single direction that carries 97.3\% of the outgoing variance. Removing that one direction restores a participation ratio of 9.76. Mid-depth rank varies 15-fold across four backbones, so it has to be measured per backbone. Retraining the decoder with high-rank input reopens the idle connections but does not improve closed-loop success. The mechanism also explains why PruneCheck must be run on each checkpoint. Supplementary Information gives the details.

\subsection{Other models}
\label{sec:beyond}

\textbf{OpenFlamingo-3B.} The gate also fails as a guide in a second gated model with the opposite structure. OpenFlamingo-3B \citep{awadalla2023openflamingo} is an image captioning model. Each of its 24 decoder blocks adds $\tanh(g)\,\mathrm{attn}(x, \mathrm{image})$ to its input $x$, so a zero gate removes the image exactly. We captioned 500 COCO images \citep{lin2014coco} and scored the captions with CIDEr \citep{vedantam2015cider}. The released model scores 83.2. The model with all gates at zero scores 5.0. The registered prediction was that the four largest-gate blocks would recover at least 0.80 of that gap. It failed: they recover 0.156. No single block costs more than 4.6 CIDEr when removed. The rank correlation between gate size and that loss is 0.05. Here the function is spread over many blocks. No four-block set we tested suffices, so a gate-based cut would remove blocks the model needs. We did not search further subsets.

\textbf{SmolVLA.} We also ran PruneCheck on SmolVLA \citep{shukor2025smolvla}, a compact VLA policy without gates. Removing all 32 of its reads leaves 0 of 40 states solved, against 14 for the unmodified policy ($p=1.2\times10^{-4}$). The public checkpoint is weak, so this is a portability check and not a third finding.

\section{Discussion}\label{sec:discussion}

\paragraph{Implications for machine learning practice} A pruning method that selects components by a score can report two extra arms next to its pruned success rate: keep one and drop all, on paired initial states with an exact test. This applies to any model with removable components and a metric that can be paired. More generally, a selection rule is a predictor. It needs the same held-out validation that a model does, on the checkpoint where it is used.

\paragraph{Guidelines for practitioners} Use at least 150 paired states and an exact paired test. Remove a connection by deleting its keys from the attention, not by zeroing its gate. Run PruneCheck on each checkpoint, because the gate was right on one release and wrong on the next. Measure latency on the target device before promising speed.

\paragraph{Cost and benefit} A cut chosen by Block Influence left LIBERO-Object at 0\% success. PruneCheck prevents such a cut for 6 to 9 GPU-hours per head. Evaluating one pruned policy on all 2,000 episodes of the four LIBERO suites takes about 40 GPU-hours. Testing all 24 connections one by one would take 25 arms, about 75 GPU-hours. The cost of PruneCheck does not grow with the number of connections, because keep one tests only the kept connection.

\paragraph{Scope of the evidence} The robot evidence comes from VLA-Adapter. It is the released VLA with a learned gate on every connection and checkpoints for all four suites in two releases. The eight heads are not one data point. They come from two releases and four suites, each with its own adapted backbone. The gate fails on three of them. Block Influence fails on five, and only one head is shared between the two sets. We do not claim that every VLA has one sufficient connection. The rank-collapse mechanism predicts that the selected connection depends on the backbone. More architectures would add more verdicts. They would not remove the need to run PruneCheck on each checkpoint. PruneCheck needs only removable components and a paired metric, as its run on OpenFlamingo-3B shows.

\paragraph{Limitations} All robot results are in simulation. A physical robot and a policy trained across suites are not covered. Most comparisons use 150 paired states. They bound large effects but do not show equivalence. The prospective test of the fallback rule covers two heads, and its candidate was suggested by the other five. Block Influence is applied here to fusion connections, which is not the setting in which the pruning methods that use it were evaluated. The Long comparison with published criteria is exploratory.

\section{Conclusion}\label{sec:conclusion}

This paper proposed PruneCheck, a closed-loop pruning framework for VLA policies. Current practice selects the components to keep with a cheap score and reports one success rate. PruneCheck uses scores only to rank candidates and selects by paired closed-loop ablation registered in advance. On eight released VLA-Adapter heads, it reduced the 24 fusion connections to one on every head. The largest gate selected a sufficient connection on five heads. Block Influence did so on two. A cut chosen by Block Influence took a policy from 100\% success to 0\%. The cut PruneCheck selected kept the shipped checkpoint at 91.8\% against 93.0\%.

We recommend that pruning studies report the keep-one and drop-all arms next to the pruned success rate. Future work should extend PruneCheck to policies trained across task suites, to other fusion designs and to physical robots. For models whose function is spread over many connections, as in OpenFlamingo-3B, a search over sets of connections is a further step.

{\small
\bibliographystyle{plainnat}
\bibliography{refs}

\begin{thebibliography}{39}
\providecommand{\natexlab}[1]{#1}
\providecommand{\url}[1]{\texttt{#1}}
\expandafter\ifx\csname urlstyle\endcsname\relax
  \providecommand{\doi}[1]{doi: #1}\else
  \providecommand{\doi}{doi: \begingroup \urlstyle{rm}\Url}\fi

\bibitem[Adebayo et~al.(2018)Adebayo, Gilmer, Muelly, Goodfellow, Hardt, and
  Kim]{adebayo2018sanity}
Julius Adebayo, Justin Gilmer, Michael Muelly, Ian Goodfellow, Moritz Hardt,
  and Been Kim.
\newblock Sanity checks for saliency maps.
\newblock In \emph{Advances in Neural Information Processing Systems},
  volume~31, pages 9505--9515, 2018.

\bibitem[Alayrac et~al.(2022)Alayrac, Donahue, Luc, Miech, Barr, Hasson, Lenc,
  Mensch, Millican, Reynolds, et~al.]{alayrac2022flamingo}
Jean-Baptiste Alayrac, Jeff Donahue, Pauline Luc, Antoine Miech, Iain Barr,
  Yana Hasson, Karel Lenc, Arthur Mensch, Katherine Millican, Malcolm Reynolds,
  et~al.
\newblock Flamingo: A visual language model for few-shot learning.
\newblock In \emph{Advances in Neural Information Processing Systems},
  volume~35, pages 23716--23736, 2022.

\bibitem[Awadalla et~al.(2023)Awadalla, Gao, Gardner, Hessel, Hanafy, Zhu,
  Marathe, Bitton, Gadre, Sagawa, Jitsev, Kornblith, Koh, Ilharco, Wortsman,
  and Schmidt]{awadalla2023openflamingo}
Anas Awadalla, Irena Gao, Josh Gardner, Jack Hessel, Yusuf Hanafy, Wanrong Zhu,
  Kalyani Marathe, Yonatan Bitton, Samir Gadre, Shiori Sagawa, Jenia Jitsev,
  Simon Kornblith, Pang~Wei Koh, Gabriel Ilharco, Mitchell Wortsman, and Ludwig
  Schmidt.
\newblock {OpenFlamingo}: An open-source framework for training large
  autoregressive vision-language models.
\newblock \emph{arXiv preprint arXiv:2308.01390}, 2023.

\bibitem[Bachlechner et~al.(2021)Bachlechner, Majumder, Mao, Cottrell, and
  McAuley]{bachlechner2021rezero}
Thomas Bachlechner, Bodhisattwa~Prasad Majumder, Henry Mao, Gary Cottrell, and
  Julian McAuley.
\newblock {ReZero} is all you need: Fast convergence at large depth.
\newblock In \emph{Proceedings of the Thirty-Seventh Conference on Uncertainty
  in Artificial Intelligence}, volume 161 of \emph{Proceedings of Machine
  Learning Research}, pages 1352--1361, 2021.

\bibitem[Baltru{\v{s}}aitis et~al.(2019)Baltru{\v{s}}aitis, Ahuja, and
  Morency]{baltrusaitis2019multimodal}
Tadas Baltru{\v{s}}aitis, Chaitanya Ahuja, and Louis-Philippe Morency.
\newblock Multimodal machine learning: A survey and taxonomy.
\newblock \emph{IEEE Transactions on Pattern Analysis and Machine
  Intelligence}, 41\penalty0 (2):\penalty0 423--443, 2019.

\bibitem[Bjorck et~al.(2025)Bjorck, Casta{\~n}eda, Cherniadev, Da, Ding, Fan,
  Fang, Fox, Hu, Huang, et~al.]{bjorck2025gr00t}
Johan Bjorck, Fernando Casta{\~n}eda, Nikita Cherniadev, Xingye Da, Runyu Ding,
  Linxi Fan, Yu~Fang, Dieter Fox, Fengyuan Hu, Spencer Huang, et~al.
\newblock {GR00T N1}: An open foundation model for generalist humanoid robots.
\newblock \emph{arXiv preprint arXiv:2503.14734}, 2025.

\bibitem[Blalock et~al.(2020)Blalock, Gonzalez~Ortiz, Frankle, and
  Guttag]{blalock2020state}
Davis Blalock, Jose~Javier Gonzalez~Ortiz, Jonathan Frankle, and John Guttag.
\newblock What is the state of neural network pruning?
\newblock In \emph{Proceedings of Machine Learning and Systems}, volume~2,
  pages 129--146, 2020.

\bibitem[Choi et~al.(2026)Choi, Lee, Park, Kim, Krishna, Fox, and
  Yu]{choi2026vlaeval}
Suhwan Choi, Yunsung Lee, Yubeen Park, Chris~Dongjoo Kim, Ranjay Krishna,
  Dieter Fox, and Youngjae Yu.
\newblock vla-eval: A unified evaluation harness for vision-language-action
  models.
\newblock \emph{arXiv preprint arXiv:2603.13966}, 2026.

\bibitem[Fei et~al.(2025)Fei, Wang, Shi, Dai, Cai, Qian, Gong, Qiu,
  et~al.]{fei2025liberoplus}
Senyu Fei, Siyin Wang, Junhao Shi, Zihao Dai, Jikun Cai, Pengfang Qian,
  Jingjing Gong, Xipeng Qiu, et~al.
\newblock {LIBERO-Plus}: In-depth robustness analysis of vision-language-action
  models.
\newblock \emph{arXiv preprint arXiv:2510.13626}, 2025.

\bibitem[Grant et~al.(2026)Grant, Zhao, and Wang]{grant2026features}
Bryce Grant, Xijia Zhao, and Peng Wang.
\newblock Not all features are created equal: A mechanistic study of
  vision-language-action models.
\newblock \emph{arXiv preprint arXiv:2603.19233}, 2026.

\bibitem[Gromov et~al.(2024)Gromov, Tirumala, Shapourian, Glorioso, and
  Roberts]{gromov2024unreasonable}
Andrey Gromov, Kushal Tirumala, Hassan Shapourian, Paolo Glorioso, and
  Daniel~A. Roberts.
\newblock The unreasonable ineffectiveness of the deeper layers.
\newblock \emph{arXiv preprint arXiv:2403.17887}, 2024.

\bibitem[Hinostroza et~al.(2026)Hinostroza, {Toro Icarte}, Devia, Carvallo,
  Herrera-Berg, Parra, and Silva]{hinostroza2026rethinking}
Cristian Hinostroza, Rodrigo {Toro Icarte}, Christ Devia, Andres Carvallo,
  Eugenio Herrera-Berg, Denis Parra, and Jorge Silva.
\newblock Rethinking layer relevance in large language models beyond cosine
  similarity.
\newblock In \emph{International Conference on Learning Representations}, 2026.

\bibitem[Jain and Wallace(2019)]{jain2019attention}
Sarthak Jain and Byron~C. Wallace.
\newblock Attention is not explanation.
\newblock In \emph{Proceedings of the 2019 Conference of the North American
  Chapter of the Association for Computational Linguistics: Human Language
  Technologies}, pages 3543--3556, 2019.

\bibitem[Kim et~al.(2024)Kim, Pertsch, Karamcheti, Xiao, Balakrishna, Nair,
  Rafailov, Foster, Lam, Sanketi, Vuong, Kollar, Burchfiel, Tedrake, Sadigh,
  Levine, Liang, and Finn]{kim2024openvla}
Moo~Jin Kim, Karl Pertsch, Siddharth Karamcheti, Ted Xiao, Ashwin Balakrishna,
  Suraj Nair, Rafael Rafailov, Ethan Foster, Grace Lam, Pannag Sanketi, Quan
  Vuong, Thomas Kollar, Benjamin Burchfiel, Russ Tedrake, Dorsa Sadigh, Sergey
  Levine, Percy Liang, and Chelsea Finn.
\newblock {OpenVLA}: An open-source vision-language-action model.
\newblock In \emph{Proceedings of the 8th Conference on Robot Learning}, volume
  270 of \emph{Proceedings of Machine Learning Research}, 2024.

\bibitem[Kim et~al.(2025)Kim, Finn, and Liang]{kim2025openvlaoft}
Moo~Jin Kim, Chelsea Finn, and Percy Liang.
\newblock Fine-tuning vision-language-action models: Optimizing speed and
  success.
\newblock \emph{arXiv preprint arXiv:2502.19645}, 2025.

\bibitem[Li et~al.(2023)Li, Li, Savarese, and Hoi]{li2023blip2}
Junnan Li, Dongxu Li, Silvio Savarese, and Steven Hoi.
\newblock {BLIP-2}: Bootstrapping language-image pre-training with frozen image
  encoders and large language models.
\newblock In \emph{Proceedings of the 40th International Conference on Machine
  Learning}, volume 202 of \emph{Proceedings of Machine Learning Research},
  pages 19730--19742, 2023.

\bibitem[Lin et~al.(2026)Lin, Chen, Liang, Xie, Liang, and
  Xu]{lin2026autopruner}
Tao Lin, Yikeng Chen, Qiwei Liang, Zinan Xie, Zhengtu Liang, and Renjing Xu.
\newblock {AutoPruner}: Agentic depth pruning for vision-language-action
  models.
\newblock \emph{arXiv preprint arXiv:2610.23880}, 2026.

\bibitem[Lin et~al.(2014)Lin, Maire, Belongie, Hays, Perona, Ramanan,
  Doll{\'a}r, and Zitnick]{lin2014coco}
Tsung-Yi Lin, Michael Maire, Serge Belongie, James Hays, Pietro Perona, Deva
  Ramanan, Piotr Doll{\'a}r, and C.~Lawrence Zitnick.
\newblock Microsoft {COCO}: Common objects in context.
\newblock In \emph{Computer Vision -- ECCV 2014}, volume 8693 of \emph{Lecture
  Notes in Computer Science}, pages 740--755. Springer, 2014.

\bibitem[Liu et~al.(2023)Liu, Zhu, Gao, Feng, Liu, Zhu, and
  Stone]{liu2023libero}
Bo~Liu, Yifeng Zhu, Chongkai Gao, Yihao Feng, Qiang Liu, Yuke Zhu, and Peter
  Stone.
\newblock {LIBERO}: Benchmarking knowledge transfer for lifelong robot
  learning.
\newblock In \emph{Advances in Neural Information Processing Systems},
  volume~36, pages 44776--44791, 2023.

\bibitem[Liu et~al.(2024)Liu, Li, Li, and Lee]{liu2024llava15}
Haotian Liu, Chunyuan Li, Yuheng Li, and Yong~Jae Lee.
\newblock Improved baselines with visual instruction tuning.
\newblock In \emph{2024 IEEE/CVF Conference on Computer Vision and Pattern
  Recognition (CVPR)}, pages 26286--26296, 2024.

\bibitem[Liu et~al.(2026)Liu, Yang, Xie, Ames, and Yue]{liu2026layermystery}
Yuxiang Liu, Lizhi Yang, Fengze Xie, Aaron Ames, and Yisong Yue.
\newblock The layer mystery of {VLA}: An information-theoretical analysis of
  {VLA} latent interface.
\newblock \emph{arXiv preprint arXiv:2609.36118}, 2026.

\bibitem[Luo et~al.(2026)Luo, Chen, Wu, Sui, Liu, Gu, Liu, Feng, Yu, Gu, Jia,
  Heng, and Zhang]{luo2026lookbefore}
Yulin Luo, Hao Chen, Zhuangzhe Wu, Bowen Sui, Jiaming Liu, Chenyang Gu,
  Zhuoyang Liu, Qiuxuan Feng, Jiale Yu, Shuo Gu, Peng Jia, Pheng-Ann Heng, and
  Shanghang Zhang.
\newblock Look before acting: Enhancing vision foundation representations for
  vision-language-action models.
\newblock \emph{arXiv preprint arXiv:2603.15618}, 2026.

\bibitem[McNemar(1947)]{mcnemar1947note}
Quinn McNemar.
\newblock Note on the sampling error of the difference between correlated
  proportions or percentages.
\newblock \emph{Psychometrika}, 12\penalty0 (2):\penalty0 153--157, 1947.

\bibitem[Men et~al.(2025)Men, Xu, Zhang, Wang, Lin, Lu, Han, and
  Chen]{men2025shortgpt}
Xin Men, Mingyu Xu, Qingyu Zhang, Bingning Wang, Hongyu Lin, Yaojie Lu, Xianpei
  Han, and Weipeng Chen.
\newblock {ShortGPT}: Layers in large language models are more redundant than
  you expect.
\newblock In \emph{Findings of the Association for Computational Linguistics:
  ACL 2025}, pages 20192--20204, 2025.

\bibitem[Michel et~al.(2019)Michel, Levy, and Neubig]{michel2019sixteen}
Paul Michel, Omer Levy, and Graham Neubig.
\newblock Are sixteen heads really better than one?
\newblock In \emph{Advances in Neural Information Processing Systems},
  volume~32, 2019.

\bibitem[Oquab et~al.(2024)Oquab, Darcet, Moutakanni, Vo, Szafraniec, Khalidov,
  Fernandez, Haziza, Massa, El-Nouby, et~al.]{oquab2024dinov2}
Maxime Oquab, Timoth{\'e}e Darcet, Th{\'e}o Moutakanni, Huy~V. Vo, Marc
  Szafraniec, Vasil Khalidov, Pierre Fernandez, Daniel Haziza, Francisco Massa,
  Alaaeldin El-Nouby, et~al.
\newblock {DINOv2}: Learning robust visual features without supervision.
\newblock \emph{Transactions on Machine Learning Research}, 2024.

\bibitem[{Physical Intelligence} et~al.(2025){Physical Intelligence}, Black,
  Brown, Darpinian, Dhabalia, Driess, Esmail, Equi, Finn, Fusai, Galliker,
  Ghosh, Groom, Hausman, Ichter, Jakubczak, Jones, Ke, LeBlanc, Levine,
  Li-Bell, Mothukuri, Nair, Pertsch, Ren, Shi, Smith, Springenberg, Stachowicz,
  Tanner, Vuong, Walke, Walling, Wang, Yu, and Zhilinsky]{intelligence2025pi05}
{Physical Intelligence}, Kevin Black, Noah Brown, James Darpinian, Karan
  Dhabalia, Danny Driess, Adnan Esmail, Michael Equi, Chelsea Finn, Niccolo
  Fusai, Manuel~Y. Galliker, Dibya Ghosh, Lachy Groom, Karol Hausman, Brian
  Ichter, Szymon Jakubczak, Tim Jones, Liyiming Ke, Devin LeBlanc, Sergey
  Levine, Adrian Li-Bell, Mohith Mothukuri, Suraj Nair, Karl Pertsch, Allen~Z.
  Ren, Lucy~Xiaoyang Shi, Laura Smith, Jost~Tobias Springenberg, Kyle
  Stachowicz, James Tanner, Quan Vuong, Homer Walke, Anna Walling, Haohuan
  Wang, Lili Yu, and Ury Zhilinsky.
\newblock $\pi_{0.5}$: A vision-language-action model with open-world
  generalization.
\newblock \emph{arXiv preprint arXiv:2504.16054}, 2025.

\bibitem[{Qwen Team}(2024)]{qwen2024qwen25}
{Qwen Team}.
\newblock {Qwen2.5} technical report.
\newblock \emph{arXiv preprint arXiv:2412.15115}, 2024.

\bibitem[Shukor et~al.(2025)Shukor, Aubakirova, Capuano, Kooijmans, Palma,
  Zouitine, Aractingi, Pascal, Russi, Marafioti, Alibert, Cord, Wolf, and
  Cadene]{shukor2025smolvla}
Mustafa Shukor, Dana Aubakirova, Francesco Capuano, Pepijn Kooijmans, Steven
  Palma, Adil Zouitine, Michel Aractingi, Caroline Pascal, Martino Russi,
  Andres Marafioti, Simon Alibert, Matthieu Cord, Thomas Wolf, and Remi Cadene.
\newblock {SmolVLA}: A vision-language-action model for affordable and
  efficient robotics.
\newblock \emph{arXiv preprint arXiv:2506.01844}, 2025.

\bibitem[Touvron et~al.(2021)Touvron, Cord, Sablayrolles, Synnaeve, and
  J{\'e}gou]{touvron2021cait}
Hugo Touvron, Matthieu Cord, Alexandre Sablayrolles, Gabriel Synnaeve, and
  Herv{\'e} J{\'e}gou.
\newblock Going deeper with image transformers.
\newblock In \emph{2021 IEEE/CVF International Conference on Computer Vision
  (ICCV)}, pages 32--42, 2021.

\bibitem[Vedantam et~al.(2015)Vedantam, Zitnick, and Parikh]{vedantam2015cider}
Ramakrishna Vedantam, C.~Lawrence Zitnick, and Devi Parikh.
\newblock {CIDEr}: Consensus-based image description evaluation.
\newblock In \emph{2015 IEEE Conference on Computer Vision and Pattern
  Recognition (CVPR)}, pages 4566--4575, 2015.

\bibitem[Voita et~al.(2019)Voita, Talbot, Moiseev, Sennrich, and
  Titov]{voita2019analyzing}
Elena Voita, David Talbot, Fedor Moiseev, Rico Sennrich, and Ivan Titov.
\newblock Analyzing multi-head self-attention: Specialized heads do the heavy
  lifting, the rest can be pruned.
\newblock In \emph{Proceedings of the 57th Annual Meeting of the Association
  for Computational Linguistics}, pages 5797--5808, 2019.

\bibitem[Wang et~al.(2020)Wang, Tran, and Feiszli]{wang2020multimodal}
Weiyao Wang, Du~Tran, and Matt Feiszli.
\newblock What makes training multi-modal classification networks hard?
\newblock In \emph{2020 IEEE/CVF Conference on Computer Vision and Pattern
  Recognition (CVPR)}, pages 12692--12702, 2020.

\bibitem[Wang et~al.(2026)Wang, Ding, Li, Cui, Ge, Tong, Song, Zhao, Zhao, Hou,
  et~al.]{wang2026vlaadapter}
Yihao Wang, Pengxiang Ding, Lingxiao Li, Can Cui, Zirui Ge, Xinyang Tong,
  Wenxuan Song, Han Zhao, Wei Zhao, Pengxu Hou, et~al.
\newblock {VLA-Adapter}: An effective paradigm for tiny-scale
  vision-language-action model.
\newblock In \emph{Proceedings of the AAAI Conference on Artificial
  Intelligence}, volume~40, pages 18638--18646, 2026.

\bibitem[Yang et~al.(2025)Yang, Wang, Wen, Zhongwei, Zou, Zhang, Wen, and
  Zhang]{yang2025efficientvla}
Yantai Yang, Yuhao Wang, Zichen Wen, Luo Zhongwei, Chang Zou, Zhipeng Zhang,
  Chuan Wen, and Linfeng Zhang.
\newblock {EfficientVLA}: Training-free acceleration and compression for
  vision-language-action models.
\newblock In \emph{Advances in Neural Information Processing Systems},
  volume~38, pages 45758--45781, 2025.

\bibitem[Yue et~al.(2024)Yue, Wang, Kang, Han, Wang, Song, Feng, and
  Huang]{yue2024deervla}
Yang Yue, Yulin Wang, Bingyi Kang, Yizeng Han, Shenzhi Wang, Shiji Song, Jiashi
  Feng, and Gao Huang.
\newblock {DeeR-VLA}: Dynamic inference of multimodal large language models for
  efficient robot execution.
\newblock In \emph{Advances in Neural Information Processing Systems},
  volume~37, pages 56619--56643, 2024.

\bibitem[Zhai et~al.(2023)Zhai, Mustafa, Kolesnikov, and Beyer]{zhai2023siglip}
Xiaohua Zhai, Basil Mustafa, Alexander Kolesnikov, and Lucas Beyer.
\newblock Sigmoid loss for language image pre-training.
\newblock In \emph{2023 IEEE/CVF International Conference on Computer Vision
  (ICCV)}, pages 11941--11952, 2023.

\bibitem[Zhang et~al.(2026)Zhang, Dong, Zhang, Heng, Chi, Dai, Du, Du, and
  Zhang]{zhang2026molevla}
Rongyu Zhang, Menghang Dong, Yuan Zhang, Liang Heng, Xiaowei Chi, Gaole Dai,
  Li~Du, Yuan Du, and Shanghang Zhang.
\newblock {MoLe-VLA}: Dynamic layer-skipping vision language action model via
  mixture-of-layers for efficient robot manipulation.
\newblock In \emph{Proceedings of the AAAI Conference on Artificial
  Intelligence}, volume~40, pages 18764--18772, 2026.

\bibitem[Zhou et~al.(2025)Zhou, Xu, Tie, Chen, Zhang, Chu, Zhou, and
  Sun]{zhou2025liberopro}
Xueyang Zhou, Yangming Xu, Guiyao Tie, Yongchao Chen, Guowen Zhang, Duanfeng
  Chu, Pan Zhou, and Lichao Sun.
\newblock {LIBERO-PRO}: Towards robust and fair evaluation of
  vision-language-action models beyond memorization.
\newblock \emph{arXiv preprint arXiv:2510.03827}, 2025.

\end{thebibliography}
}

\clearpage
\appendix
\section*{Supplementary Information}
\setcounter{section}{0}\setcounter{table}{0}\setcounter{figure}{0}
\renewcommand{\thesection}{S\arabic{section}}
\renewcommand{\thetable}{S\arabic{table}}
\renewcommand{\thefigure}{S\arabic{figure}}
\renewcommand{\theHsection}{S\arabic{section}}
\renewcommand{\theHtable}{S\arabic{table}}
\renewcommand{\theHfigure}{S\arabic{figure}}
This document holds the details that the main text summarizes. Section, table and figure numbers with the prefix S refer to this document.

\section{Target policy: details}

VLA-Adapter \citep{wang2026vlaadapter} pairs a Qwen2.5-0.5B language-model backbone \citep{qwen2024qwen25} with a 24-block action decoder (the corresponding figure of the main texta). Two image encoders, DINOv2 \citep{oquab2024dinov2} and SigLIP \citep{zhai2023siglip}, turn the camera images into patch tokens for the backbone. Each decoder block has one attention operation with one softmax. In this softmax, block $l$ attends to three groups of tokens. The first two are its own action tokens and the proprioceptive tokens, which encode the robot's own state. The third is the 512 patch tokens that leave backbone layer $l+1$. Blocks are numbered from 0 and layers from 1, so block 23 reads the last layer. A learned gate, $\tanh(g_l)$, multiplies the attention scores of the patch tokens. We call each such read a \emph{connection}. The policy has 24 connections, one per decoder block.

The authors release eight trained checkpoints, which we call heads. Four are first-release heads, one per LIBERO suite: Spatial, Object, Goal and Long. Four are heads of a later revision, called Pro, on the same suites. The Pro decoder uses a revised block class and its own adapted backbones. The backbone is adapted per suite: 370 of its 982 tensors differ between any two first-release heads. Each head is therefore a complete policy with its own backbone and decoder.

Removing a connection means deleting block $l$'s patch tokens from the keys of its softmax. On Pro blocks, this is done by passing an empty visual key slice. Zeroing the gate would not remove the connection. With a zero gate, every patch token still receives the same non-zero attention weight in the shared softmax. The block then receives the mean of the patch tokens. Each removal wrapper is checked at load time. With every connection kept, it reproduces the original forward pass bit for bit. The package's implementation of the first-release block agrees with the original to $6.6\times10^{-7}$ in 32-bit floating point (fp32). With one connection removed, the output is finite and changes.

\section{Screens: calibration data}

Gate census, selectivity, rank and Block Influence need forward passes only and no labels. Each takes minutes per checkpoint on a T4. Block Influence uses 50 observations per head: five initial states of each task, after the standard settling steps. $R$ needs 20 batches of 8 demonstration frames.

\section{Step-by-step protocol}

Table~\ref{tab:protocol} lists the steps of PruneCheck with their recorded cost. Step~6 is the fallback rule of the main text.

\begin{table*}[!tp]
\caption{\textbf{PruneCheck, step by step.} Costs are recorded runs on one Tesla T4. No screen overrides the ablation.}
\label{tab:protocol}
\small
\begin{tabularx}{\linewidth}{@{}c>{\raggedright\arraybackslash}p{3.6cm}>{\raggedright\arraybackslash}p{2.2cm}L@{}}
\toprule
Step & Measurement & Cost & How to read it \\
\midrule
1 & Gate census, input rank (PR), Block Influence per connection & forward passes, minutes & PR near one marks a candidate to exclude, never a verdict. The largest gate or score is a candidate, not an answer. \\
2 & Attention selectivity per connection & one forward, minutes & Uniform attention does not mean disuse. Two heads rely on a connection with selectivity $\le 10^{-6}$. \\
3 & Offline recovery $R$ (the corresponding equation of the main text) & minutes per head & Low $R$ flags that the largest gate is not sufficient. It ranks which heads to roll out first. \\
4 & Keep one and drop all, 150 paired states, exact McNemar & about 3\,h per arm & The only step that decides. \\
5 & If keep one matches the unmodified head & none & Prune to that connection. Adapt briefly if desired. \\
6 & If keep one costs as much as drop all & 3\,h per candidate & Do not prune to the largest gate. Ablate other single candidates, the deepest first. \\
\bottomrule
\end{tabularx}
\end{table*}

\section{Experimental record}

Every closed-loop experiment was registered before it ran. Each registration fixed three things. The first is a numeric prediction. The second is a falsifier, the result that would refute the prediction. The third is a set of instrument conditions. These are checks that the measurement itself works, such as reproducing the original forward pass. A failed condition is recorded as ``not tested'', never as a success.

Rollouts use LIBERO's fixed initial states and $256\times256$ images. They run in 16-bit floating point (fp16) on Tesla T4 GPUs. An fp32 check over 109 paired episodes does not distinguish fp32 from fp16 ($-2.8$ points, $p=0.51$). The comparison of the main text uses a second software environment. It runs the VLA-Adapter authors' released code with their package versions, in fp16, with software rendering by OSMesa. With every connection kept, its removal wrapper reproduces the original forward pass bit for bit on all eight heads.

\section{Failures of the cheap screens: full text}

We now compare each screen of the main text with the closed-loop verdict (the corresponding table of the main text).

\textbf{Gate census.} Gate magnitude as a ranking fails on three of eight heads, as shown above. An absolute reading fails too. On first-release Spatial and Goal, the connection that suffices alone has a gate of 0.0045 and 0.0016. Both values are below 0.01, the threshold we registered for calling a gate open. A reader would call such a connection inactive.

\textbf{Attention selectivity.} Selectivity is near zero on the same two heads. Their needed connection averages its 512 tokens to six decimal places. Its selectivity is $1.0\times10^{-6}$ on Spatial and $1.0\times10^{-7}$ on Goal. Yet removing it together with the other 23 costs 76.7 and 72.0 points. Block 23 was not removed on its own. On Goal, it also ranks below another connection, with selectivity $2.4\times10^{-7}$. Uniform attention is therefore not evidence of disuse. A connection that delivers only the mean of its tokens can still carry the policy.

\textbf{Input rank.} Input rank was registered as a screen on all eight heads. The prediction was $\mathrm{PR} > 2$ at the largest-gate block exactly where that block suffices. Six of eight heads fell on the registered side, and two did not. Pro-Spatial and Pro-Goal have rank 8.45 and 5.11 at a block that is worth nothing alone. Rank can exclude a connection, as on Pro-Object with 1.19. It cannot confirm one.

\textbf{Offline recovery.} Offline recovery separated the eight heads correctly. Where the largest gate suffices, $R$ ranges from 0.974 to 0.997. Where it does not, $R$ ranges from 0.005 to 0.048. The separation is 0.93. Offline recovery is itself an ablation, scored on loss instead of success. It was registered with the eight verdicts already known. This is therefore a retrospective test of agreement on one family's training frames. We use it to order rollouts, not to replace them.

In summary, none of the screens can stand in for the closed-loop test. Each one is useful only to decide which connection to test first.

\section{Registered comparison: reproduction condition}

This comparison used the second software environment (the main text). A registered condition therefore required each new unmodified run to land within 5 successes of the count recorded in the first environment. The condition held on five heads. The new runs solved 144 against 141 on Spatial, 150 against 150 on Object, 149 against 146 on Goal, 150 against 149 on Pro-Spatial, and 148 against 143 on Pro-Goal. It failed on Long, with 144 against 138. The Long comparison is therefore reported as exploratory. Pro-Long and Pro-Object have no recorded count.

\section{Prospective test of the protocol: full text}

Step~6 of Table~\ref{tab:protocol} handles the failure case: when the largest gate fails, ablate the deepest block. We wrote this step after the Pro failures, on the evidence of the five heads where block 23 suffices. We then registered it as a prediction on the two failing heads where block 23 had not been tried. The prediction was that keeping only block 23 on Pro-Spatial and Pro-Goal would change success by no worse than $-8$ points. The test used the same code, instrument and 150 paired states. As an instrument condition, twenty episodes of each unmodified head were re-run. They had to reproduce the recorded outcomes.

The prediction held on both heads. On Pro-Spatial, keeping only block 23 solves 150 of 150 states. The unmodified head solves 149 ($+0.7$ points, McNemar $p=1.0$). On Pro-Goal, keeping only block 23 solves 148 of 150. The unmodified head solves 143 ($+3.3$, $p=0.125$). Keeping the largest gate solves only 115 and 16 on these heads. The replicate episodes agreed 20 of 20 on both heads. On these two heads, one connection therefore suffices. It is the same block as on the other five, and the gate points away from it.

Pro-Object was not part of this test. On Pro-Object, keeping its two largest gates was also no better than none, at $-92.7$ points. Block 23 was rolled out on Pro-Object as the C3 pick of the registered comparison in the main text. It solves 150 of 150 states. The unmodified head solves 147 ($+2.0$ points, McNemar $p=0.25$).

One connection therefore suffices on all eight heads, and it is block 23 on each. Step~6 thus turns a failed screen into a recovered cut. The price is one extra arm, about 3 GPU-hours.

\section{Pruning the shipped checkpoint: full text}

The deployment question is what the cut costs. the corresponding table of the main text collects the registered measurements on the Long head. We pruned the shipped checkpoint at inference to block 23, with weights untouched. We evaluated it on all 500 initial states of LIBERO-10, the benchmark's name for the Long suite. The pruned checkpoint scores $459/500 = 91.8\%$. The unmodified checkpoint scores $465/500 = 93.0\%$. The change is $-1.2$ points. The two arms disagree on 36 episodes, 21 in one direction and 15 in the other, and $p=0.41$. The registered bound of $-4.0$ points held. A cost of up to about 3.5 points is not excluded. The cut removes 82.2\% of the decoder's multiply-accumulate operations, its basic arithmetic. The floor is one connection, not zero. With all 24 connections dropped, the policy falls to 4.0\%.

Brief adaptation narrows the gap. We fine-tuned only the action decoder of the pruned policy, with the backbone frozen. Training ran for 4,000 steps at learning rate $2\times10^{-5}$ on the suite's demonstrations and took 47 GPU-minutes. The adapted policy scores 92.7\% against the released 92.0\% on 150 paired states. Three training seeds give 92.7\%, 92.0\% and 90.7\%. None differs significantly from the released head. The same recipe gives 96.0\% on Spatial against 94.0\%. It gives 98.7\% on Object against 100.0\% and 96.7\% on Goal against 97.3\%. Two of these registered predictions did not hold. The per-run floor of $-1.0$ points was crossed by one Long seed and by Object. Each crossed it by a third of a point, which is half an episode. Both are recorded as inconclusive.

The saving is in arithmetic, not in time. We measured latency on a T4 at batch size one. The call of the pruned action decoder falls from 20.5 to 16.5\,ms, a 19.6\% saving. The whole action query, from images to action chunk, falls from 146.9 to 141.3\,ms in the medians. The median of the 20 paired savings is 1.8\%. These savings range from $-19.8\%$ to $+13.5\%$, which is within noise. The backbone dominates the call. The registered prediction of at least a 50\% speed-up of the action decoder was refuted. On this hardware, the case for pruning is a smaller decoder, not a faster robot. The case for PruneCheck is that the cut it selects does not harm the policy. The cut the gate suggests can destroy it.

\section{Mechanism behind the single connection: full text}

The single connection has a measurable cause: most connections receive tokens that carry almost no variation. The tokens that the backbone hands to mid-depth connections have a participation ratio near one in all four first-release checkpoints. the corresponding table of the main text reports the deep block. Over the 20 mid-depth blocks, the per-checkpoint median is 1.08 to 1.14. These tokens form a tight cluster. Attending over them returns nearly the same vector whatever the query asks.

\begin{figure*}[!tp]
\centering
\includegraphics[width=0.8\linewidth]{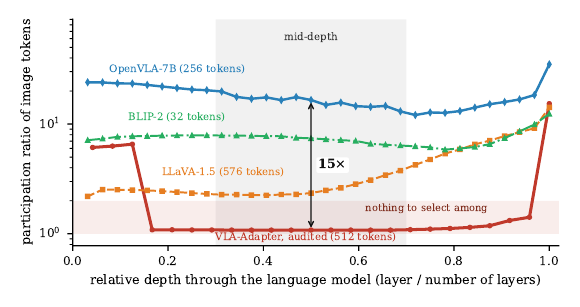}
\caption{\textbf{The rank of the image tokens is a property of the backbone.} Participation ratio of image tokens per layer of the language model, on three LIBERO-Long tasks. Two curves are VLA backbones: VLA-Adapter, the target policy, and OpenVLA-7B \citep{kim2024openvla}. Two are vision--language models: LLaVA-1.5 \citep{liu2024llava15} and BLIP-2 \citep{li2023blip2}. The ratio cannot exceed the token count given in the labels. The target backbone has twice OpenVLA-7B's tokens yet 15-fold lower mid-depth rank. The VLA-Adapter curve uses the first-release Long head over 10 policy calls. Its last layer reaches 15.3, and 17.63 over the 50 frames of the corresponding table of the main text. The other curves use one agent-view frame per task.}
\label{fig:rank}
\end{figure*}

We localized this collapse inside the backbone. Each of the 24 backbone layers writes to the hidden state twice, once from its attention and once from its MLP. We deleted these 48 writes one at a time. The fourth layer's MLP write injects a single direction, with its own participation ratio of 1.002. This direction carries 97.3\% of the outgoing variance. Removing that one direction restores a participation ratio of 9.76. The collapse is a property of the backbone, not of the architecture class. Mid-depth rank varies 15-fold across four backbones (Fig.~\ref{fig:rank}). It therefore has to be measured per backbone.

Repairing the collapse does not improve the policy. Retraining the decoder with high-rank input and repaired gate dynamics reopens the idle connections. A reopened interface does not do better in closed loop. The corrected 24-connection head scores $-8.7$ points against its matched twin with the released interface (exact McNemar $p=0.053$). The mechanism also explains why PruneCheck must be run on each checkpoint. Rank collapse depends on the backbone, so a verdict on one checkpoint does not transfer to another.

\section{Other models: full text}

\textbf{OpenFlamingo-3B.} Gate magnitude also fails as a guide in a second gated model, which has the opposite structure. OpenFlamingo-3B \citep{awadalla2023openflamingo} is an image captioning model. It gates a cross-attention block at each of 24 decoder blocks as
\begin{equation*}
x + \tanh(g)\,\mathrm{attn}(x, \mathrm{image}).
\end{equation*}
A zero gate therefore removes the image exactly. Its gates are large, from 0.011 to 0.392. We captioned 500 COCO images \citep{lin2014coco} with greedy decoding. Each prompt held four example captions. We scored each caption with CIDEr \citep{vedantam2015cider}, a standard measure of caption quality. The released model scores 83.2. The model with all gates at zero scores 5.0. The registered prediction was that the four largest-gate blocks would recover at least 0.80 of that gap. It failed: they recover 0.156. No single block costs more than 4.6 CIDEr when removed. The rank correlation between gate size and that loss is 0.05. The four largest gates do beat the four smallest, with 0.148 more recovery. Here the gate ranks blocks coarsely. What fails is the cut it would suggest. No four-block set we tested suffices, and no single block is necessary. A gate-based pruning would therefore remove blocks the model needs. We did not search further subsets. The ablation shows that the gate's cut is unsafe. It does not show which smaller set would be safe.

\textbf{SmolVLA.} We also ran PruneCheck on SmolVLA \citep{shukor2025smolvla}, a compact VLA policy, using the public LIBERO checkpoint. This tests whether the harness carries over to a policy without gates and on a different software stack. The action expert of SmolVLA reads the VLM prefix at each of 32 layers. On 40 paired LIBERO-10 states, the harness's unmodified policy solved 14. The reference evaluator of the lerobot library solved 13 on the same states ($-2.5$ points, $p=1.0$). The harness's rollout is therefore not distinguishable from the reference at this size. Removing all 32 reads leaves 0 of 40 ($p=1.2\times10^{-4}$). Keeping only the 16 joint-attention reads also leaves 0 of 40, as does keeping only the 16 cross-attention reads. These two arms are exploratory, because a half-ablated model is off its training distribution. The checkpoint solves only 32.5 to 40\% of LIBERO-10 under both pipelines. This is well below the figure reported for the suite. SmolVLA is therefore a portability check, not a third finding. Together, the two models show that the failure of the gate is not specific to one robot policy. They also show that PruneCheck runs on interfaces without gates.

\section{PruneCheck harness and registered record}

This section describes the software that runs PruneCheck and lists every registered outcome. Its purpose is to show that the instrument itself was tested before its results were used.

\textbf{Implementation.} PruneCheck was implemented as a Python package, \texttt{vla\_audit}, with three parts. An adapter names the connections of a model family. It switches all but a chosen set off for the duration of one run and restores the model afterwards. An engine runs every arm on the same paired items. It appends each result to disk and resumes after a crash. A report computes exact McNemar or paired bootstrap tests. It flags every connection where a screen and the ablation disagree. The package was tested with 51 unit tests that run without a model or a GPU. A toy policy reproduces the flagging in one command. The comparison of the main text used a separate implementation of the VLA-Adapter adapter and runner. It passed the same bit-exactness condition on all eight heads.

\textbf{Hardware.} Every arm is paired and resumable, so PruneCheck runs on a single commodity GPU. The comparison of the main text ran on two T4 GPUs of a preemptible instance and resumed after a restart.

\textbf{Validation of the harness.} The harness was validated under the same rules as the experiments. Each validation run was registered in advance and is listed in Table~\ref{tab:validation}. The VLA-Adapter adapter reproduced all 450 recorded closed-loop outcomes of three arms on 150 states. The OpenFlamingo adapter reproduced all 15\,000 recorded captions of 30 arms on 500 images. These references are our own scripts on the same GPU class. They test the implementation, not the method independently. The SmolVLA adapter had no recorded runs, so we checked it against the reference evaluator of the lerobot library instead.

Five of the eight harness runs held. One missed: the SmolVLA run predicted at least 16 of 40 for the unmodified policy and got 14. Two runs failed the harness's own instrument conditions and are recorded as not tested. One was an fp32 equality registered at exactly 0, which a correct implementation reached to $6.6\times10^{-7}$. The other was a precision condition written without measuring what the reference evaluator runs.

\textbf{Tests of the method.} Three tests of the method were registered. The offline screen is a retrospective test. Step~6 and the comparison with published criteria are prospective tests. All three held. The comparison's reproduction condition held on five heads and failed on one.

\begin{table*}[!tp]
\caption{\textbf{Every registered validation outcome, in the order run.} ``Not tested'' means a registered instrument condition failed as written. The prediction was then neither confirmed nor refuted. No run was repeated or dropped.}
\label{tab:validation}
\small
\begin{tabularx}{\linewidth}{@{}>{\raggedright\arraybackslash}p{3.9cm}LL>{\raggedright\arraybackslash}p{1.5cm}@{}}
\toprule
Run & Registered prediction & Observed & Verdict \\
\midrule
OpenFlamingo, 4 arms $\times$ 40 images & all 160 captions as recorded & 160 of 160 & held \\
VLA-Adapter, 2 arms $\times$ 20 states & full $\ge 16$, none $\le 3$; fp32 forward equal at 0 & 17 and 0; equal to $6.6\times10^{-7}$ & not tested \\
OpenFlamingo probes, 16 images & selectivity and rank within $10^{-6}$ & equal to 0 & held \\
VLA-Adapter, 3 arms $\times$ 150 states & $\ge 147$ of 150 per arm as recorded & 450 of 450 & held \\
OpenFlamingo, 30 arms $\times$ 500 images & $\ge 14\,850$ of 15\,000 as recorded & 15\,000 & held \\
SmolVLA, 2 arms $\times$ 10 states & full $\ge 5$, none $\le 2$ & 5 and 0 & held \\
SmolVLA, 4 arms $\times$ 40 states & full $\ge 16$, none $\le 4$, cost $\ge 12$ & 14, 0, 14 & not held \\
SmolVLA, reference evaluator & 8 to 20 solved; fp32 & 13; reference runs bf16/fp32 & not tested \\
Method: offline recovery, 8 heads, retrospective & 8 of 8 on the right side, separation $\ge 0.3$ & 8 of 8, 0.93 & held \\
Method: Step 6, keep block 23, 2 heads, prospective & $\ge -8$ points on each & $+0.7$ and $+3.3$ & held \\
Second environment, unmodified arms & within 5 successes of the recorded count & 5 of 6 heads with a record & held on 5, not on Long \\
Method: Block Influence picks, 8 heads, prospective & C1 picks block 0 or 3 on $\ge 6$ heads; some criterion fails on some head & block 0 on 6; fails on 7 heads & held \\
\bottomrule
\end{tabularx}
\end{table*}

\section{Cost of a wrong cut and cost of PruneCheck}

Table~\ref{tab:value} sets the cost of a wrong cut against the cost of PruneCheck.

\begin{table*}[!tp]
\caption{\textbf{What a wrong cut costs and what PruneCheck costs.} Changes are paired against the unmodified head. The worst case is taken over the heads where each rule could be evaluated. GPU-hours are measured on one Tesla T4.}
\label{tab:value}
\small
\begin{tabularx}{\linewidth}{@{}>{\raggedright\arraybackslash}p{5.2cm}L@{}}
\toprule
Decision or cost & Measured outcome \\
\midrule
Keep the largest gate & up to $-90.7$ points; wrong on 3 of 8 heads \\
Keep the Block Influence pick, C1 & where it picks block 0, $-68.7$ to $-100.0$ points on all 5 heads evaluated; LIBERO-Object falls from 100\% to 0\% \\
Keep the PruneCheck pick, block 23 & $-1.2$ points over 500 episodes on the shipped checkpoint; 82.2\% fewer decoder multiply-accumulates \\
PruneCheck screens & minutes per checkpoint, no labels \\
PruneCheck decision & 6 to 9 GPU-hours per checkpoint \\
Full evaluation of a pruned policy, four suites & about 40 GPU-hours for 2,000 episodes \\
\bottomrule
\end{tabularx}
\end{table*}

\end{document}